\documentclass[10pt,logo,copyright]{nvidia_report}
\usepackage[authoryear,sort&compress,round]{natbib}
\setcitestyle{numbers}
\usepackage{colortbl}
\usepackage{xspace}
\usepackage{enumitem}
\usepackage{amsmath,amssymb}
\usepackage{booktabs}
\usepackage{longtable}
\usepackage{array}
\usepackage{algpseudocode}
\usepackage{algorithm}
\usepackage{titletoc}
\usepackage{minted}
\usepackage{listings}
\usepackage{xltabular}
\usepackage[nameinlink]{cleveref}
\usepackage{caption}
\usepackage{wrapfig2}

\newcommand{\para}[1]{\paragraph{#1}\looseness=-1}
\newcommand{\acronym}[0]{\mbox{\textsc{RoboTTT}}\xspace}
\newcommand{\fullname}[0]{\mbox{\emph{Test-Time-Training Robot Policies}}\xspace}
\newcommand{\webpage}[0]{\href{https://research.nvidia.com/labs/gear/robottt/}{\textbf{\texttt{research.nvidia.com/labs/gear/robottt}}}}

\newcommand{\groot}[0]{GR00T N1.7\xspace}
\newcommand{\grootHist}[0]{GR00T N1.7 Hist.\xspace}
\newcommand{\gdn}[0]{GDN\xspace}

\renewcommand{\para}[1]{\par\noindent
  {\normalsize\bfseries #1}\hspace{1em}\looseness=-1\ignorespaces}
  
\title{\acronym: Context Scaling for Robot Policies}
\author{
\parbox{\textwidth}{
\raggedright
{\small\bfseries
Yunfan Jiang$^{1,2}$,
Yevgen Chebotar$^{1}$,
Ruijie Zheng$^{1}$,
Fengyuan Hu$^{1}$,
Yunhao Ge$^{1}$}\\
{\small\bfseries
Jimmy Wu$^{1}$,
Tianyuan Dai$^{1,3}$,
Scott Reed$^{1}$,
Li Fei-Fei$^{2,\dagger}$,
Yuke Zhu$^{1,3,\dagger}$,
Linxi ``Jim'' Fan$^{1,\dagger}$}\\
{\footnotesize
$^{1}$NVIDIA \quad
$^{2}$Stanford University \quad
$^{3}$The University of Texas at Austin \quad
$^{\dagger}$Equal advising}
}\\
{\webpage}
}
\keywords{Long-Context Policies, Test-Time Training, Robot Foundation Models}

\begin{document}

\maketitle

\begin{center}
    \includegraphics[width=\textwidth]{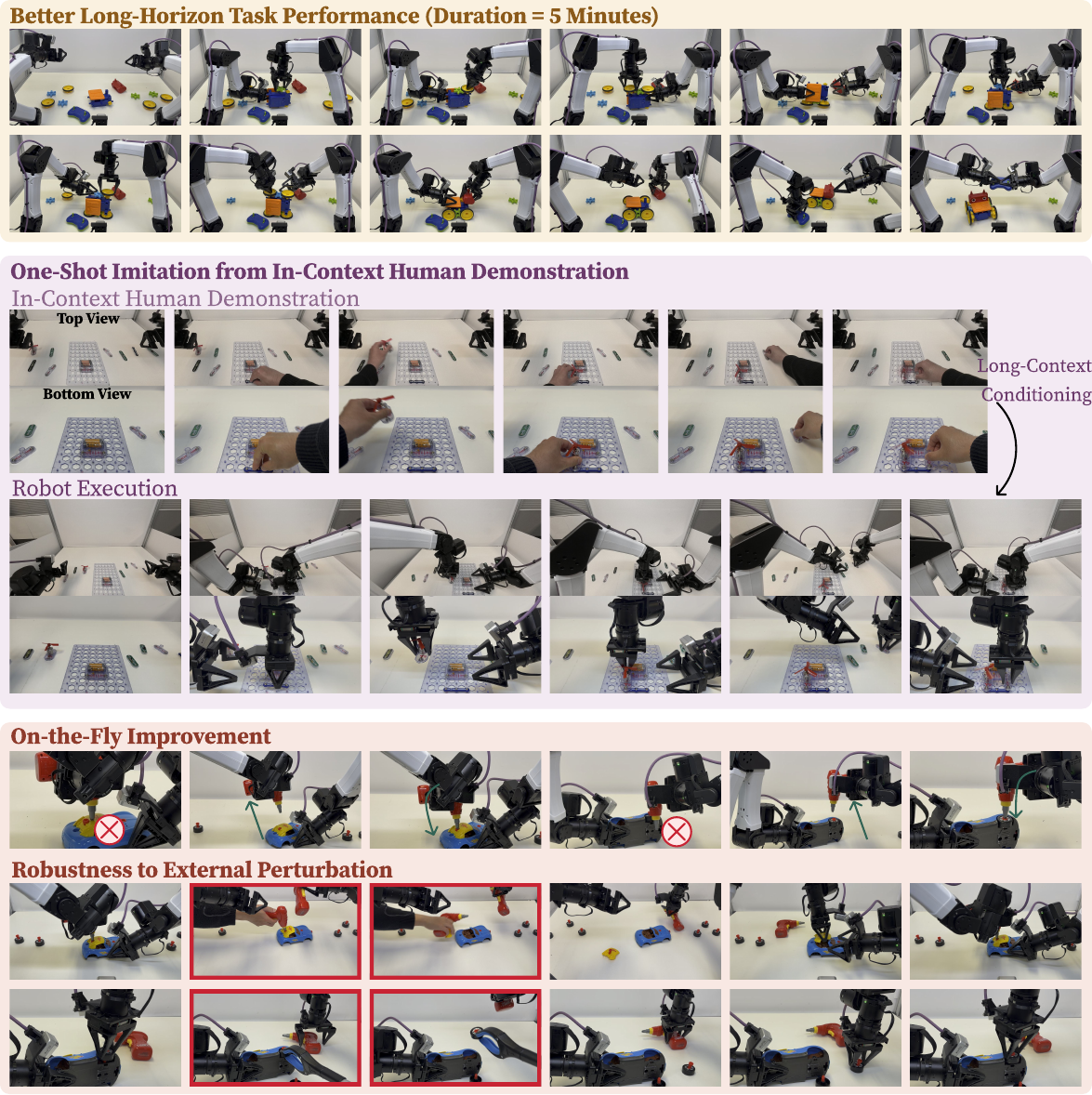}
\end{center}
\vspace{-0.4cm}
\begingroup
\small
\refstepcounter{figure}\label{fig:pull}
\looseness=-1 Figure~\thefigure: \textbf{\acronym, a long-context visuomotor policy that integrates Test-Time Training (TTT) into robot foundation models, with context scaled to 8K timesteps.} \acronym exhibits capabilities such as one-shot in-context imitation from human videos and on-the-fly policy improvement.
\par
\endgroup

\clearpage
\begin{abstract}
Recent robot foundation models operate with single-step or short-history visuomotor context.
We introduce \fullname (\acronym), a robot model and training recipe that scale visuomotor context to 8K timesteps, three orders of magnitude beyond state-of-the-art policies, without growing inference latency.
At this context length, we unlock new robot capabilities: one-shot in-context imitation from human video demonstrations, on-the-fly policy improvement, robustness to perturbations, and stronger performance on multi-stage, long-horizon tasks.
We also observe, for the first time, steady gains in closed-loop performance as pretraining context length scales.
At its core, \acronym integrates Test-Time Training into robot foundation models such as Vision-Language-Action policies, yielding a sequence model whose recurrent state consists of \emph{fast weights}, parameters updated by gradient descent during both training and inference, compressing histories into weight space and retrieving contextual information for long-context conditioning.
To scale training context length, the recipe combines \emph{sequence action forcing} with \emph{truncated backpropagation through time}.
On challenging real-robot manipulation tasks, \acronym improves overall performance by 87\% over the single-step context baseline and fully completes a five-minute, ten-stage assembly task, which no baseline ever does.
\acronym trained with 8K-timestep context outperforms the same model pretrained with 1K timesteps by 62\%, suggesting context length as a new scaling axis for robot foundation models.
Videos are available at \webpage.

\end{abstract}
\abscontent

\section{Introduction}
\label{sec:introduction}
Most state-of-the-art robot foundation models operate with single-step or short-history visuomotor context~\citep{nvidia2025gr00t,black2410pi0,pi05,kim2024openvla,shukor2025smolvla0,yuan2026fast0wam0,kim2026cosmos,brohan2022rt1,brohan2023rt2,team2024octo,DBLP:conf/iclr/LiuWLTCWX0025,li2025unified,ye2026world}.
In contrast, context length has become an important scaling axis for large language models~\citep{brown2020language,DBLP:conf/icml/DingZZXSX0Y24,DBLP:conf/iclr/PengQFS24,DBLP:conf/iclr/XiaoTCHL24,hsieh2024ruler0,liu2024lost}.
While longer-term reasoning in robotics can be delegated to external memory banks~\citep{torne2026mem0,mark2026bpp0}, long visuomotor context remains important for capabilities such as one-shot in-context imitation from human video demonstrations~\citep{duan2017one}, on-the-fly improvement from a robot's own deployment history~\citep{DBLP:conf/iclr/LaskinWOPSSSHFB23}, and stronger closed-loop performance on multi-stage, long-horizon tasks.
This raises a natural question: \emph{how can we build visuomotor policies that learn from and exploit arbitrarily long contexts?}

To answer this question, we introduce \fullname (\emph{\acronym}, Fig.~\ref{fig:pull}), a robot model and training recipe that scale visuomotor context to 8K timesteps (dubbed \acronym-8K), three orders of magnitude beyond state-of-the-art robot foundation models~\citep{nvidia2025gr00t,black2410pi0,ye2026world}, without growing inference latency.
At this context length, \acronym exhibits new robot capabilities.
Through long-context conditioning, it performs one-shot imitation from a single in-context human video demonstration, succeeding in 6 of 10 trials while baseline methods fail entirely.
It also exhibits on-the-fly policy improvement, performing 36\% better than the same model not trained for this capability.
Under external perturbations, it succeeds in 83\% of trials versus 53\% for the best short-context baseline.
On multi-stage, long-horizon tasks, it improves overall task performance by 87\% over the single-step context baseline and fully completes an assembly task lasting over \textbf{five minutes} and spanning \textbf{ten stages}, which no baseline ever does.
Finally, we show for the first time that scaling pretraining context length yields steady gains in closed-loop performance: \acronym-8K achieves a 63\% higher task completion score than the same model pretrained with 1K-timestep context and outperforms the best short-context baseline by 57\%, suggesting context length as a new scaling axis for robot foundation models.

At its core, \acronym integrates Test-Time Training (TTT)~\citep{sun2024learning,zhang2025test0time} into robot foundation models such as Vision-Language-Action (VLA) policies~\citep{nvidia2025gr00t}.
\acronym is a sequence model whose recurrent state consists of \emph{fast weights}~\citep{schlag2021linear}: unlike \emph{slow weights}, which are frozen at inference, fast weights are updated by gradient descent during both training and inference.
This design addresses the three challenges of long-context visuomotor policies: encoding long histories with sufficient capacity, exploiting the conditioned context~\citep{dai2026robomme}, and keeping inference cost constant in context length.
First, a \emph{fast model} (e.g., an MLP) parameterized by fast weights offers greater capacity than the vector-valued states of recurrent neural networks.
Second, training the fast model during deployment retains salient features and discards redundant ones in the dense, repetitive streams of robot observations and actions.
Third, propagating fast weights over time keeps inference cost constant, whereas Transformer inference grows with history even with a KV cache.
Crucially, by learning during deployment, \acronym enables new forms of in-context adaptation and policy improvement.
For example, it conditions on a human video demonstrating a new task configuration to achieve one-shot imitation.
Through \emph{DAgger Distillation}, a training procedure that distills DAgger-style~\citep{dagger} failure-to-correction mappings into fast weights, it learns to improve on the fly.
To scale training context length, our recipe combines \emph{sequence action forcing} with \emph{truncated backpropagation through time} (TBPTT), allowing context to grow without increasing GPU memory.
Videos are available at \webpage.

\section{Preliminaries}
\label{sec:preliminary}
\para{Test-Time Training Mechanism} Test-Time Training (TTT)~\citep{sun2024learning,zhang2025test0time} introduces \emph{fast weights} that are updated during both training and inference to dynamically model contextual information. This contrasts with \emph{slow weights} (i.e., model parameters), which are updated only during training and remain frozen during inference.
Formally, consider a sequence of $d$-dimensional tokens $X$ and its query, key, and value sequences $Q, K, V$, induced by projection matrices $\theta_Q, \theta_K, \theta_V$, with $Q_t, K_t, V_t$ denoting the projections at timestep $t$.
The fast weights $W$ parameterize a small neural network $f_W(\cdot): \mathbb{R}^d \rightarrow \mathbb{R}^d$, for instance a linear layer or an MLP.
At timestep $t$, the fast weights are updated to associate $K$ with its corresponding value projection $V$ through
\begin{equation}
\label{eq:fw_update}
    W_t \leftarrow W_{t-1} - \eta \nabla_W \mathcal{L}_\mathrm{FW}\left(f_{W_{t-1}}\left(K_t\right), V_t\right),
\end{equation}
where $\mathcal{L}_\mathrm{FW}(\hat{v}, v) = \Vert \hat{v} - v \Vert^2$ is typically a mean squared error and $\eta$ denotes the (learnable) learning rate. The updated fast weights then compute the output $O_t$ in the \emph{apply} step via
\begin{equation}
\label{eq:ttt_output}
    O_t = f_{W_t}(Q_t).
\end{equation}
This ``update then apply'' operation occurs during both training and inference. Intuitively, the update step encodes contextual information into the parameter space of the fast model $f_W$, and the apply step retrieves it for the downstream prediction. The projection matrices $\theta_Q, \theta_K, \theta_V$ and the fast weight initialization $W_0$ are learned with the outer task loss, such that the history mechanism is optimized for the task at hand. At inference, TTT thus compresses previously processed tokens into its fast weights, whereas standard full attention retains all previous keys and values in memory and attends over them at each step.

\para{Robot Sequence Models}
In this work, we consider robot policies that condition on their rollout history, also known as robot sequence models~\citep{DBLP:journals/tmlr/ReedZPCNBGSKSEBREHCHVBF22,jiang2022vima,team2024octo,fu2024context}.
Concretely, a robot trajectory $\xi = \left\{\left( o_t, q_t, A_t \right) \right\}^{T}_{t=1}$ consists of image $o_t$, proprioception $q_t$, and action chunk $A_t$ tuples; we omit the language modality for simplicity. We learn policies $\pi(A_t \vert \xi_{<t}, o_t, q_t)$ that condition on the history $\xi_{<t}$ before timestep $t$ and the current observation. For long-context policies, we aim to scale the context length $\vert \xi_{<t} \vert$.

\begin{figure}[t]
\centering
\includegraphics[width=\textwidth]{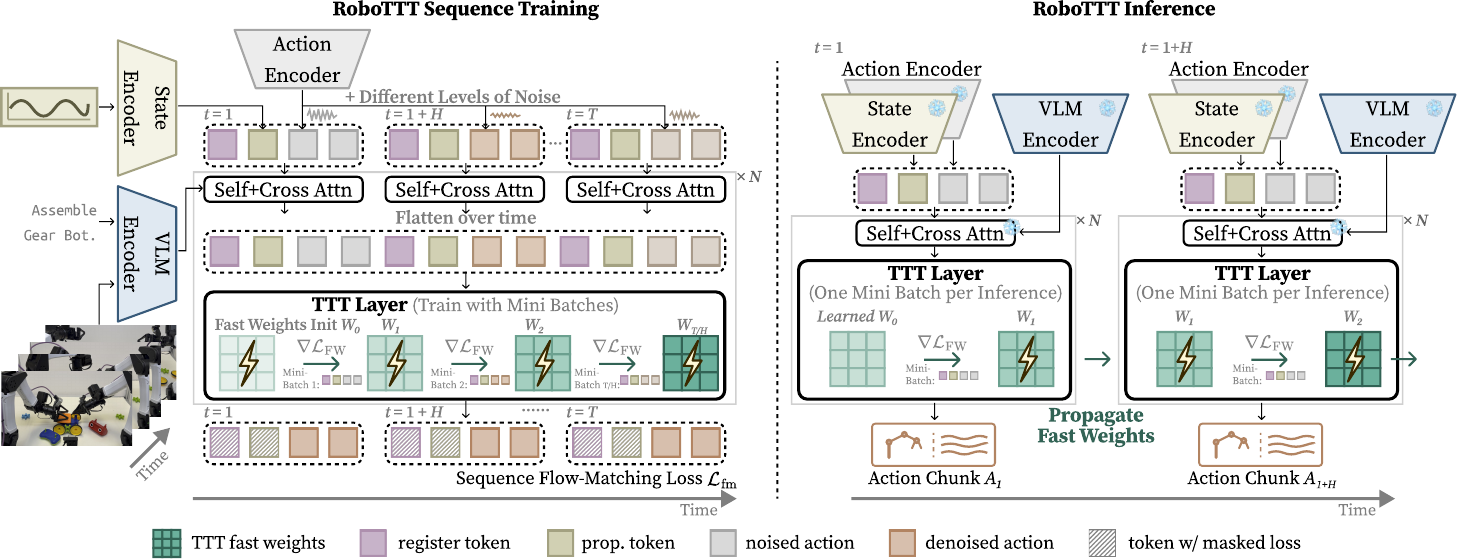}
\vspace{-0.6cm}
\caption{\textbf{\acronym model architecture, training, and inference.} TTT layers are added after the attention layers in the DiT action head: attention operates within each timestep, while TTT layers operate across timesteps. Training uses a sequence flow-matching loss with \emph{sequence action forcing}, sampling the noise level independently per action chunk. Inference starts from the learned initialization $W_0$, updating fast weights on each observation and propagating them forward.}
\label{fig:model_arch}
\vspace{-0.2cm}
\end{figure}

\section{Method: Test-Time-Training Robot Policies}
\label{sec:method}
In this section, we introduce \fullname (\acronym), a robot model and training recipe for learning over long-context robot trajectories.
We first describe the model architecture and how we integrate TTT into modern robot foundation models, then a training recipe that combines \emph{sequence action forcing} with \emph{truncated backpropagation through time} (TBPTT) to scale training context length.
We then describe how \acronym enables long-context conditioning, unlocking new forms of in-context adaptation and policy improvement:
one-shot imitation from in-context human video demonstrations, and \emph{DAgger Distillation}, a meta-learning method that teaches the policy to improve on-the-fly by distilling the DAgger-style~\cite{dagger} correction process, mapping suboptimal robot actions to human corrections, into its fast weights.
Finally, we share implementation details.

\subsection{Model Architecture}
\acronym integrates TTT layers into robot foundation models, such as VLA models, while preserving compatibility with the underlying architecture, and thus applies to a broad range of backbones. In this paper, we instantiate \acronym on top of GR00T~N1.7~\citep{nvidia2025gr00t}.

\para{Test-Time Training for Robot Actions}\label{sec:model_arch_ttt} As shown in Fig.~\ref{fig:model_arch}, \acronym consists of a vision-language model (VLM) backbone and a Diffusion Transformer (DiT)~\citep{peebles2022scalable} action head with TTT layers.
At timestep $t$, it predicts an $H$-step action chunk $A_t = [a_t, \ldots, a_{t+H-1}]$.
We add TTT layers after the self- and cross-attention layers, so that attention processes single-step information while TTT layers process information across the time dimension.
Concretely, the input to \acronym's DiT is a robot trajectory spanning $T$ timesteps, $[R_1, \Phi_1, q_1, \tilde{A}_{1}, \ldots, R_T, \Phi_{T}, q_{T}, \tilde{A}_{T}]$, where $\Phi_t$ are the vision-language (VL) tokens output by the VLM, $q_t$ is the encoded proprioception token, $\tilde{A}_{t}$ are the noised action tokens, and $R_t$ are $N$ learned register tokens~\citep{jaegle2021perceiver0,DBLP:conf/iclr/DarcetOMB24} prepended at each timestep that attend to all other tokens.
Attention layers operate on the single-step tokens $R_t$, $q_t$, and $\tilde{A}_{t}$, and cross-attend to the VL tokens $\Phi_t$ of that timestep.
The per-timestep attention outputs are then concatenated along the time dimension, $X = [R_1, q_1, \tilde{A}_{1}, \ldots, R_T, q_{T}, \tilde{A}_{T}]$, and passed through the TTT layers for the fast weight update (Eq.~\ref{eq:fw_update}) and output (Eq.~\ref{eq:ttt_output}).
We avoid passing the VL tokens $\Phi$ through TTT layers directly for computational efficiency, relying instead on the smaller number ($N = 16$) of register tokens $R$ to carry VL information across time.

\begin{wrapfigure}[13]{r}{0.4\textwidth}
\vspace{-12pt}
    \includegraphics[width=0.4\textwidth]{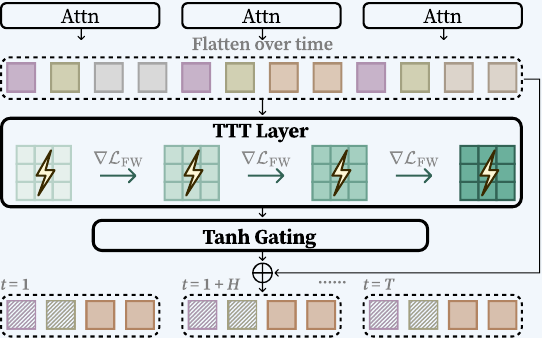}
\caption{\textbf{Computation flow with $\tanh$ gating.} The TTT output is weighted by the learned gate $\tanh(\alpha)$ before being added to the attention output.}
    \label{fig:gating}
\end{wrapfigure}

\para{Gating for Preserving Pretrained Capabilities} To retain the knowledge of the pretrained VLA model, \acronym is initialized from the base model weights and adopts a learned $\tanh$ gating mechanism~\citep{alayrac2022flamingo0} that keeps the contribution of TTT small at the start of training.
Concretely, for each DiT layer, we learn $\alpha \in \mathbb{R}^d$, initialized to near zero (0.001), and gate the TTT contribution through
\begin{equation}
    O = \tanh(\alpha) \odot O_{\mathrm{TTT}} + O_{\mathrm{attn}},
\end{equation}
where $O_{\mathrm{TTT}}$ is the TTT layer output from Eq.~\ref{eq:ttt_output} and $O_{\mathrm{attn}}$ is the attention layer output.
In this way, \acronym learns to adjust the contribution of the TTT layers without overwhelming the computation of the pretrained model.

\subsection{\acronym Sequence Training}
We train \acronym on robot trajectory sequences so that fast weights are updated within each training sequence, learning both a suitable fast weight initialization and its update dynamics.
The TTT projection matrices $\theta_Q, \theta_K, \theta_V$ and the fast weight initialization $W_0$ are learned as part of the model parameters.
Specifically, given a training sequence, we run TTT over it in the inner loop with the fast weight loss $\mathcal{L}_{\mathrm{FW}}$ (Sec.~\ref{sec:preliminary}), compute the outer task loss at every timestep, and optimize the full model on the averaged loss.
In this way, the projection matrices are learned directly from the outer task gradient, and $W_0$ is meta-learned through gradients of gradients~\citep{tandon2025end0to0end,finn2017model0agnostic}, tailoring the fast weight updates to robot trajectories.

Concretely, our dataset $\mathcal{D} = \{\xi^{(i)}\}_{i=1}^{N}$ contains $N$ trajectories, each a sequence of language, image, proprioception, and action-chunk tuples $\xi = \{(l, o_t, q_t, A_t)\}_{t=1}^{T}$, reintroducing the language instruction $l$, which is shared across the trajectory. Each training sequence is a full trajectory or a contiguous sub-trajectory up to a maximum context length.
Denoting the per-step flow-matching objective as $\ell_t$, the sequence loss over a trajectory $\xi$ given fast weight initialization $W_0$ is
\begin{equation}
    \mathcal{L}_{\mathrm{fm}}\left(\xi; W_0\right) = \frac{1}{T}\sum_{t=1}^{T}\ell_t\left(\left(l, o_t, q_t, A_t \right); W_{t - 1} \right),
\end{equation}
where $W_{t-1}$ is the fast weight state entering timestep $t$, updated to $W_t$ inside the TTT layers (Eq.~\ref{eq:fw_update}).
Each optimization step then takes a gradient step w.r.t. $\mathcal{L}_{\mathrm{fm}}$, updating both regular model weights and the fast weight initialization.

\para{Sequence Action Forcing}
\acronym is trained with a flow-matching objective for actions: the DiT action head learns to denoise $\tilde{A}_t = A^{\tau}_t = \tau A_t + (1-\tau) \epsilon$, where $\tau \in [0, 1]$ is the flow-matching timestep and $\epsilon \sim \mathcal{N}(\mathbf{0}, \mathbf{I})$ is the sampled noise.
In sequence training, we find it necessary to sample the noise level \emph{independently} for each action chunk in the sequence, a technique we call \emph{sequence action forcing}.
Without it, training is unstable, possibly because sharing one noise level across the sequence (full-sequence diffusion) makes entire sequences uniformly easy (low noise) or uniformly hard (high noise) to learn, echoing the findings of \citet{chen2024diffusion}.

To summarize, denoting the DiT action head as $v_\theta$ and the timestep-$t$ tuple as $\xi_t = (l, o_t, q_t, A_t)$, we train \acronym with sequence action forcing by minimizing
\begin{equation}
\label{eq:seq_loss}
    \mathcal{L}_{\mathrm{fm}}\left(\xi; W_0\right) = \frac{1}{T}\sum_{t=1}^{T}\ell_t\left(\xi_t; W_{t-1} \right) = \frac{1}{T}\sum_{t=1}^{T} \mathbb{E}_{\tau_t, \epsilon} \left[ \left\Vert v_\theta\left(\Phi_t, A_t^{\tau_t}, q_t; W_{t-1} \right) - \left(A_t - \epsilon\right) \right\Vert^2\right],
\end{equation}
where each $\tau_t$ is sampled independently as $\tau_t = s(1-u)$, $u \sim \mathrm{Beta}(1.5, 1)$, $s = 0.999$.

\begin{wrapfigure}[13]{r}{0.4\textwidth}
\vspace{-8pt}
    \includegraphics[width=0.4\textwidth]{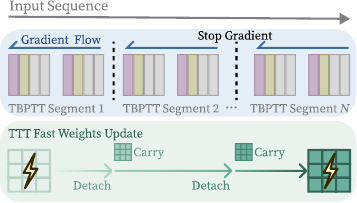}
    \caption{\textbf{TBPTT.} Gradients are truncated at segment boundaries; fast weights carry over, so TTT continues over the entire sequence.}
    \label{fig:tbptt}
\end{wrapfigure}

\para{Truncated Backpropagation Through Time}
Training on long sequences with full backpropagation through time (BPTT) stores activations for every timestep, so GPU memory grows with sequence length.
We instead adopt truncated backpropagation through time (TBPTT): the input sequence is divided into segments, and gradients flow only within each segment (Fig.~\ref{fig:tbptt}).
Crucially, the fast weights are carried over across segment boundaries, so TTT continues over the entire sequence, while their gradients are detached at the boundaries.
GPU memory is thus determined by the segment length rather than the total sequence length, allowing arbitrarily long training contexts under a fixed memory budget.
Note that the fast weight initialization $W_0$ still receives gradients through the first segment, whose updates originate directly from $W_0$.

\para{Inference}
As illustrated in Fig.~\ref{fig:model_arch}, \acronym starts each rollout from the learned initialization $W_0$, updates the fast weights on the current observation, and propagates them to the next timestep.
At each timestep, action chunks are generated with $k$-step denoising.

\subsection{Effective Learning from Context}
\acronym decouples fast weight updates from slow weight updates: by masking the flow-matching loss on selected timesteps, those timesteps serve as pure context, updating the fast weights without providing an imitation target.
This flexibility lets \acronym learn from heterogeneous contexts, such as human video demonstrations or the robot's own suboptimal rollouts.

\begin{figure}[t]
\centering
\includegraphics[width=\textwidth]{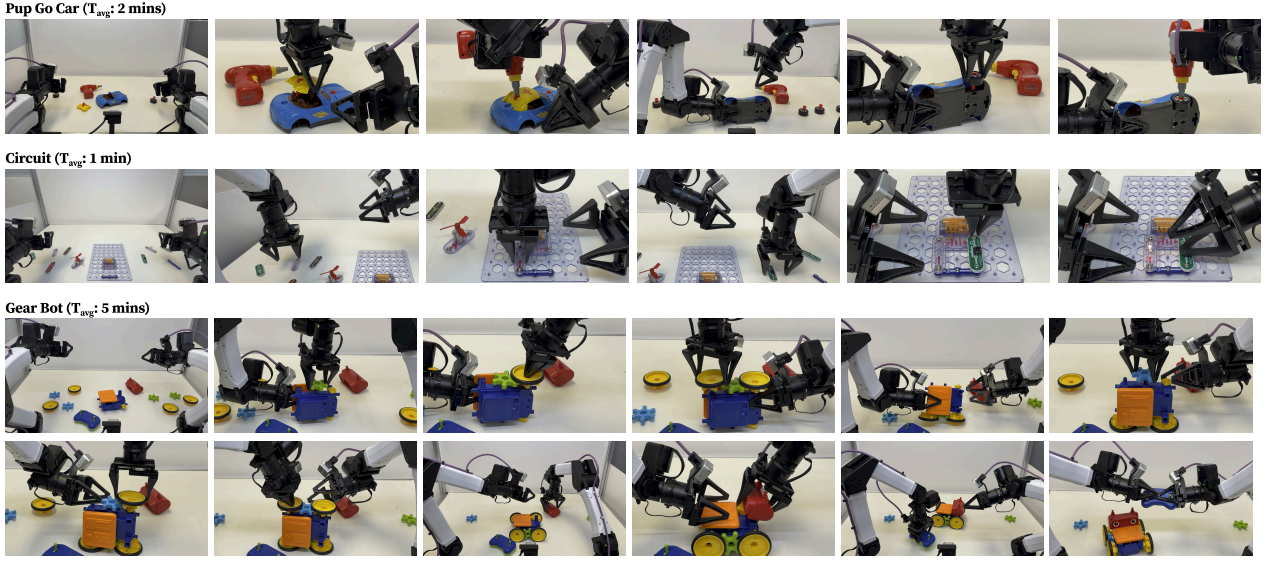}
\caption{\textbf{Evaluation tasks.} Three long-horizon assembly tasks on a YAM bimanual setup; each row shows one rollout. \textbf{Top, Pup Go Car:} toy vehicle assembly, 2-minute average episodes. \textbf{Middle, Circuit:} target configuration specified by a language prompt or a one-shot human video demonstration; 1-minute episodes. \textbf{Bottom, Gear Bot:} ten-stage assembly spanning five minutes, our longest-horizon task.}
\label{fig:tasks}
\end{figure}

\para{Imitation from In-Context Video Demonstrations}\label{sec:learning_from_context}
We pair human video demonstration sequences $\xi^{\mathrm{video}}$ with robot trajectories $\xi^{\mathrm{robot}}$ of the same task: the video sequence updates the fast weights only (its flow-matching loss is masked), while the action loss is computed on the robot trajectory conditioned on the updated fast weights.
Trained on such pairs, the model learns to extract task information from the in-context video; at test time, conditioning on a single human video of an unseen task configuration yields one-shot imitation.

\begin{wrapfigure}[11]{r}{0.4\textwidth}
\vspace{-10pt}
    \includegraphics[width=0.4\textwidth]{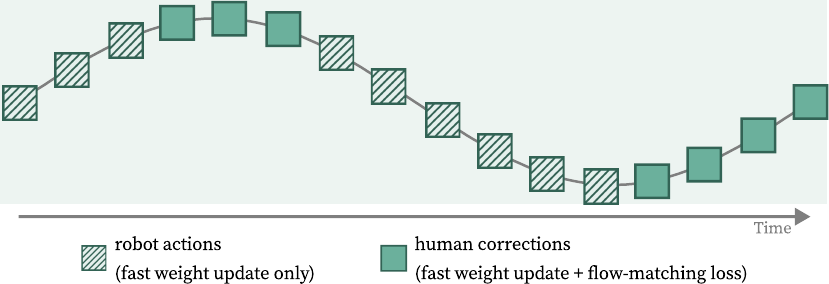}
\caption{\textbf{DAgger Distillation.} During sequence training, all executed actions update the fast weights, but the flow-matching loss is computed only on human corrections.}    \label{fig:dagger_distill}
\end{wrapfigure}

\para{DAgger Distillation}\label{sec:dd}
Consider a robot rollout with human corrections collected as in DAgger~\citep{dagger}: whenever the robot makes a mistake, a human operator intervenes with corrective actions, yielding a trajectory $\xi^{\mathrm{DAgger}} = \{(l, o_t, q_t, A_t)\}_{t=1}^{T}$ in which each executed action chunk $A_t$ is either a robot action $A^{\mathrm{R}}_t$ or a human correction $A^{\mathrm{H}}_t$.
Such interleaved rollouts trace a natural pattern of online policy improvement.
Standard DAgger fine-tunes on the human corrections and discards the suboptimal robot actions; yet it is precisely these actions that reveal what failure each correction responds to.
\acronym instead uses both, in asymmetric roles: during sequence training, the fast weights are updated on the full interaction history, not only the executed corrections but also the suboptimal robot actions themselves, while the flow-matching loss is masked to the human corrections only.

We call this procedure \emph{DAgger Distillation} (Fig.~\ref{fig:dagger_distill}).
This asymmetry, failures as context and corrections as targets, is precisely how the human's failure-to-correction mapping is distilled into the fast weights: the model learns to produce corrections in response to failures, rather than to imitate corrections in isolation, making fuller use of the same collected data.
We view it as an instantiation of Algorithm Distillation~\citep{DBLP:conf/iclr/LaskinWOPSSSHFB23} in robotics: the improvement process induced by human interventions is distilled into the policy's fast-weight adaptation.
At test time, the model performs such corrections online, without human intervention; its own corrections then enter the history and are absorbed into the fast weights, exactly as the human corrections were during training.
As we show in Sec.~\ref{sec:experiments}, DAgger Distillation yields stronger failure recovery and higher task performance than standard DAgger training.

\subsection{Implementation Details}
We instantiate \acronym on pretrained GR00T~N1.7~\citep{nvidia2025gr00t}, adding a TTT layer to each of its 16 DiT layers; each fast model is a two-layer MLP.
We pretrain on a mixture of tabletop bimanual robot data and egocentric human data~\citep{zheng2026egoscale0}, gradually increasing the pretraining context length up to the target (e.g., 8K timesteps for \acronym-8K), for 30K steps on 16 NVIDIA GB200 GPUs.
We then post-train on each downstream task at 1K context length for 20K steps.
Further details are in Appendix~\ref{appendix:sec:policy_training_details}.

\begin{figure}[t]
\centering
\includegraphics[width=0.8\textwidth]{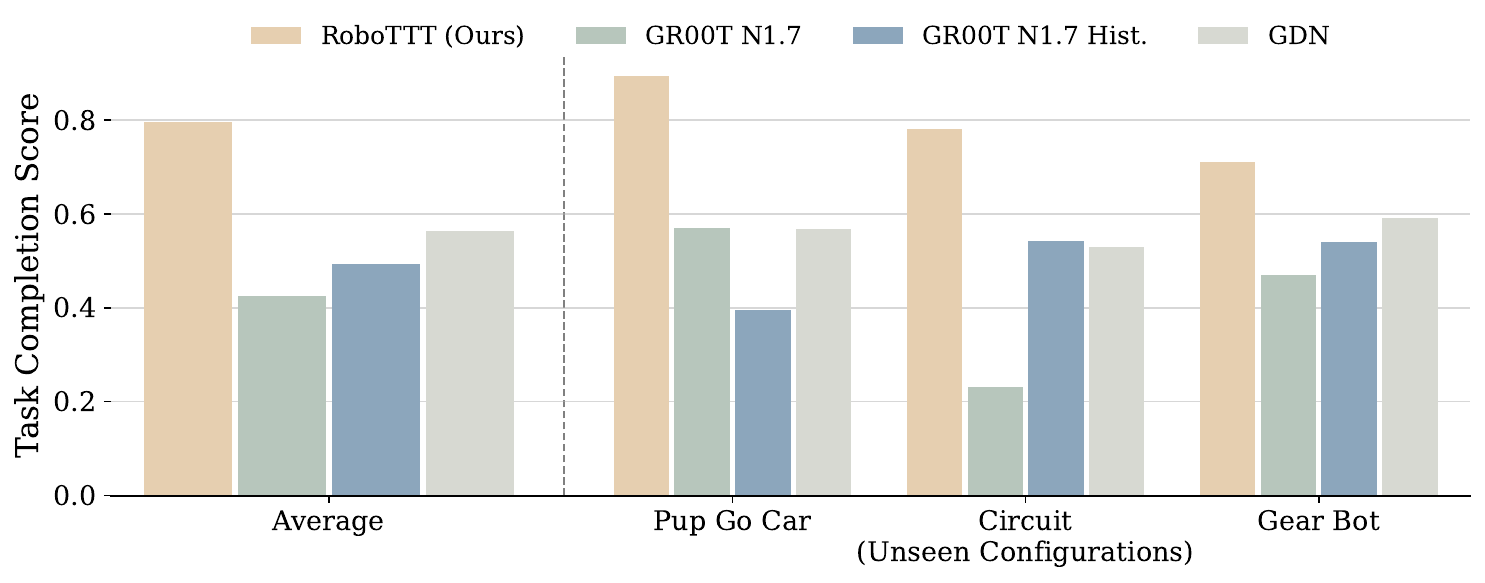}
\caption{\textbf{Main evaluation: task completion scores on three assembly tasks.} Scores are rubric-based, reported in percent; higher is better.}
\label{fig:main_exp_bar}
\end{figure}

\section{Experiments}
\label{sec:experiments}
\para{Experiment Settings} We evaluate on three long-horizon, multi-stage assembly tasks requiring bimanual manipulation and dexterity (Fig.~\ref{fig:tasks}): Pup Go Car, Circuit, and Gear Bot.
All experiments use a YAM bimanual setup with four RGB cameras: top, bottom, left wrist, and right wrist.
For each task, we collect 8, 6, and 5 hours of real-robot data, with average episode lengths of 2 minutes, 1 minute, and 5 minutes, respectively.
The Circuit task has 80 assembly configurations, varying components, assembly order, and number of pieces; the target configuration is specified through the language prompt (or a one-shot human video demonstration). We train on 20 configurations and test on the remaining 60.

We compare \acronym against three baselines: \textbf{\groot}~\citep{nvidia2025gr00t} with single-step context; \textbf{\grootHist}, GR00T N1.7 with one history frame; and \textbf{\gdn}, in which the TTT layers of \acronym are replaced with Gated DeltaNet layers~\citep{yang2024gated}, a linear-complexity recurrent memory that updates its state without test-time gradient descent.
All methods are post-trained on the same task data: sequence models use a 1K-timestep context, and non-sequence models are trained to a matched compute budget.
Each policy is evaluated for 20 trials (10 for Gear Bot due to its substantially longer horizon) across varied configurations.
We report the number of fully successful trials and a rubric-based task completion score normalized to $[0,1]$ (reported in percent).
Hyperparameters and score rubrics are in Appendix~\ref{appendix:sec:experiment_settings}.

\para{\acronym consistently outperforms baselines on dexterous, long-horizon tasks.}
As shown in Fig.~\ref{fig:main_exp_bar} and Table~\ref{tab:main_exp_result}, \acronym achieves an average task completion score of 79\%, 87\% higher than the single-step context baseline \groot (42\%) and 41\% higher than the best baseline \gdn (56\%), with the most fully successful trials on every task.
Notably, on Gear Bot, which requires five minutes on average for full completion, \acronym achieves full successes (2 of 10) while no baseline ever does.
Qualitatively, \acronym excels in three aspects.
First, it tracks task progress: in multi-stage assembly, visually similar stages cause state aliasing, leading baselines to perform wrong motions or skip stages, whereas the fast weights, updated on the fly, retain salient features of the history and disambiguate the current stage.
Second, it exhibits \emph{strategic recovery}: during roof drilling on Pup Go Car, if the drill misses the roof screw, \acronym raises the arm, re-aligns, and re-attempts, while baselines proceed to the next stage as if the previous one had succeeded.
Third, it is more precise in fine-grained stages such as inserting and snapping circuit components; we attribute this to long context mitigating partial observability, as past observations of the object of interest inform actions when it is currently occluded.
The relevant observation window is difficult to specify a priori; \acronym instead learns what to retain, suggesting that with a sufficiently expressive sequence model, the use of long context can be learned rather than hand-designed.

\begin{wraptable}{r}{0.4\textwidth}
\vspace{-8pt}
    \centering
        \resizebox{0.4\textwidth}{!}{
\begin{tabular}{@{}rccc@{}}
\toprule
\textbf{Method}  & \textbf{Pup Go Car} & \textbf{Circuit} & \textbf{Gear Bot} \\ \midrule
RoboTTT          & \textbf{9 / 20}     & \textbf{13 / 20} & \textbf{2 / 10}   \\
GR00T N1.7       & 3 / 20              & 3 / 20           & 0 / 10            \\
GR00T N1.7 Hist. & 0 / 20              & 8 / 20           & 0 / 10            \\
GDN              & 3 / 20              & 8 / 20           & 0 / 10            \\ \bottomrule
\end{tabular}
}
\caption{\textbf{Main evaluation: fully successful trials on three assembly tasks}, out of 20 per task (10 for Gear Bot). \acronym is the only method with full successes on Gear Bot.}
\label{tab:main_exp_result}
\vspace{5pt}
\end{wraptable}

Can history alone match \acronym?
Naively concatenating past observations does not reliably help: on Pup Go Car, \grootHist scores 39.5\% versus 57\% for its no-history counterpart \groot, as appended histories can introduce spurious correlations and leave the robot temporally out of distribution at inference.
\gdn, which compresses history into a recurrent state, improves over \groot on Circuit and Gear Bot but not Pup Go Car.
Notably, both \gdn and \acronym maintain fixed-size states; the difference is the update rule.
We hypothesize that the gated delta rule, a linear associative update without test-time gradient descent, struggles to extract structure from dense, repetitive robot streams across thousands of timesteps, whereas \acronym's nonlinear fast model, updated by gradient descent at test time, is the more expressive compressor.

\begin{figure}[t]
\centering
\includegraphics[width=0.8\textwidth]{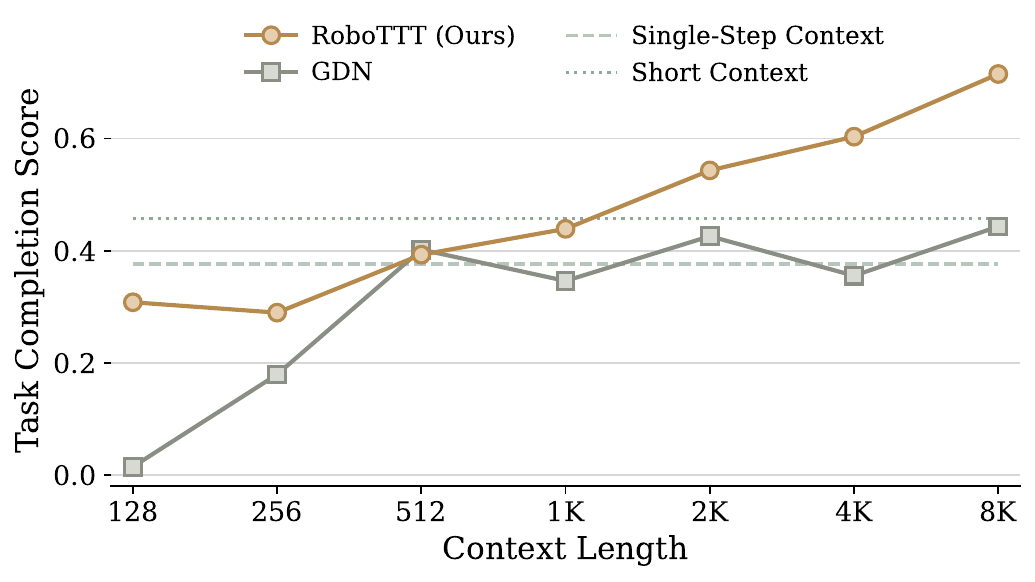}
\vspace{-10pt}
\caption{\textbf{Closed-loop performance scales with pretraining context length.} Average task completion score across the three tasks as pretraining context length grows from 128 to 8K timesteps. \acronym improves steadily, surpassing the best short-context baseline from 1K onward with no sign of saturation; \gdn does not benefit from longer context. \emph{Single-Step Context} and \emph{Short Context} denote \groot and \grootHist. All evaluations in this figure predate the DAgger training used for Pup Go Car in the main results.}
    \label{fig:ctx_scaling}
\end{figure}

\para{Scaling pretraining context length yields steady gains in closed-loop performance.}
We pretrain both \acronym and \gdn at context lengths from 128 timesteps (a few seconds) to 8K (over four minutes), then post-train and evaluate closed-loop performance on all three tasks (Fig.~\ref{fig:ctx_scaling}).
As references, \emph{Single-Step Context} denotes \groot, conditioned on the current observation only, and \emph{Short Context} denotes \grootHist, with one additional history frame.
\acronym exhibits a clear scaling trend: closed-loop performance increases steadily with pretraining context length, reaching 71.5\% at 8K, 63\% higher than the same model pretrained at 1K (43.9\%) and 57\% higher than the best short-context baseline \grootHist (45.6\%), with no sign of saturation.
\gdn shows no such trend.
We attribute the difference to the two update rules: \acronym's fast weights are updated by gradient descent, and the outer loss meta-learns their initialization $W_0$ and update dynamics, which longer training sequences shape over more update steps; \gdn's linear associative state admits no such meta-learning.
Below 1K, \acronym remains competitive (still outperforming \gdn at matched lengths) but falls short of its longer-context variants.
We attribute this to the rollout horizon exceeding the training context: 1K timesteps is about half a minute, shorter than our shortest task episode, so at inference the fast weights are updated far beyond the window seen in training, and positional embeddings extend to unseen positions.

\begin{figure}[t]
\centering
\includegraphics[width=\textwidth]{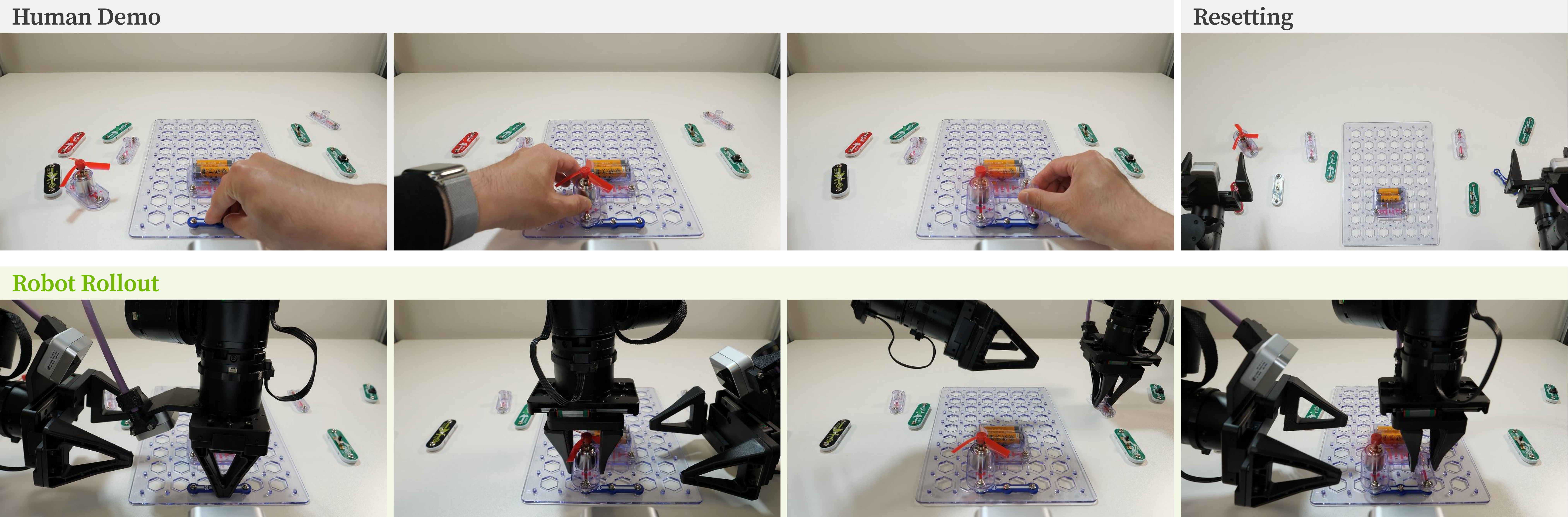}
\vspace{-0.6cm}
\caption{\textbf{One-shot imitation from an in-context human video.} Continuous footage: a human demonstrates an unseen configuration (frames 1--3), the scene is reset (frame 4), and \acronym reproduces the assembly (row 2). The prompt is identical across configurations, so the target is identifiable only from the video.}\label{fig:circuit_one_shot_rollout}
\vspace{-0.2cm}
\end{figure}

\para{Long-context conditioning unlocks one-shot imitation and perturbation robustness.}
We first evaluate one-shot imitation.
For the Circuit task, we collect additional human demonstration videos in which the robot stays idle and a human assembles the circuit by hand.
For each assembly configuration in the training set, we collect 5--20 human videos with varied initial layouts.
During training, we sample a human video and a robot trajectory of the same configuration and concatenate them into a single training sequence, masking the flow-matching loss on the video portion (Sec.~\ref{sec:learning_from_context}).
We use the same task prompt, ``assemble circuit,'' for all configurations, so the target configuration is identifiable only from the in-context video.
Since \groot cannot condition on context and human videos far exceed \grootHist's history window, we compare against \gdn.

\begin{wraptable}[7]{r}{0.4\textwidth}
\vspace{-10pt}
    \centering
    \resizebox{0.4\textwidth}{!}{
\begin{tabular}{@{}rcc@{}}
\toprule
\textbf{} & \textbf{Task Completion Score} & \textbf{Number of Successful Rollouts} \\ \midrule
RoboTTT   & \textbf{65\%}                  & \textbf{6 / 10}                        \\
GDN       & 33\%                           & 0 / 10                                 \\ \bottomrule
\end{tabular}
}
\vspace{-5pt}
    \caption{\textbf{One-shot imitation on Circuit.} Task completion score and fully successful trials, conditioning on one in-context human video of an unseen configuration.}
    \label{table:circuit_one_shot}
    \vspace{-5pt}
\end{wraptable}

As shown in Table~\ref{table:circuit_one_shot}, \acronym follows the in-context demonstration (Fig.~\ref{fig:circuit_one_shot_rollout}) and achieves six successful assemblies out of ten trials, while \gdn fails entirely, picking up wrong components or assembling in the wrong order.
This suggests that although recurrent-memory policies can encode contextual information, they struggle to use it; \acronym instead queries a fast model updated by gradient descent on the history, retrieving the demonstrated configuration when needed.

\begin{wraptable}[9]{r}{0.4\textwidth}
\vspace{-10pt}
    \centering
    \resizebox{0.4\textwidth}{!}{
\begin{tabular}{@{}rcc@{}}
\toprule
\textbf{Method}  & \textbf{Roof Perturbation} & \textbf{Tire Perturbation} \\ \midrule
RoboTTT          & \textbf{15 / 20}           & \textbf{18 / 20}           \\
GR00T N1.7       & 10 / 20                    & 11 / 20                    \\
GR00T N1.7 Hist. & 3 / 20                     & 5 / 20                     \\
GDN              & 13 / 20                    & \textbf{18 / 20}                 \\ \bottomrule
\end{tabular}
}
\vspace{-5pt}
\caption{\textbf{Robustness to external perturbations.} Each entry reports successful recoveries, i.e., where the policy reassembled the removed part, out of 20 per condition.}
\label{table:perturbation_results}
\end{wraptable}

We next evaluate perturbation robustness: beyond conditioning across episodes, \acronym also conditions \emph{within} an episode.
On Pup Go Car, a human removes the yellow roof after the robot installs it, or removes a tire after insertion; a policy that conditions on its own rollout should return to the pre-perturbation stage and reinstall the part.
We collect 30 minutes of perturbation data and co-train it with the task data.
As shown in Table~\ref{table:perturbation_results}, all methods exhibit some robustness, likely from this co-training, but the long-context methods react successfully more often: \acronym recovers in 15/20 roof-perturbation trials versus 13/20 for \gdn and at most 10/20 for the short-context baselines, and both \acronym and \gdn recover in 18/20 tire trials.
This supports the view that visuomotor context improves within-episode conditioning; example recoveries are shown in Fig.~\ref{fig:pull}.

\begin{figure}[t]
\centering
\includegraphics[width=0.8\textwidth]{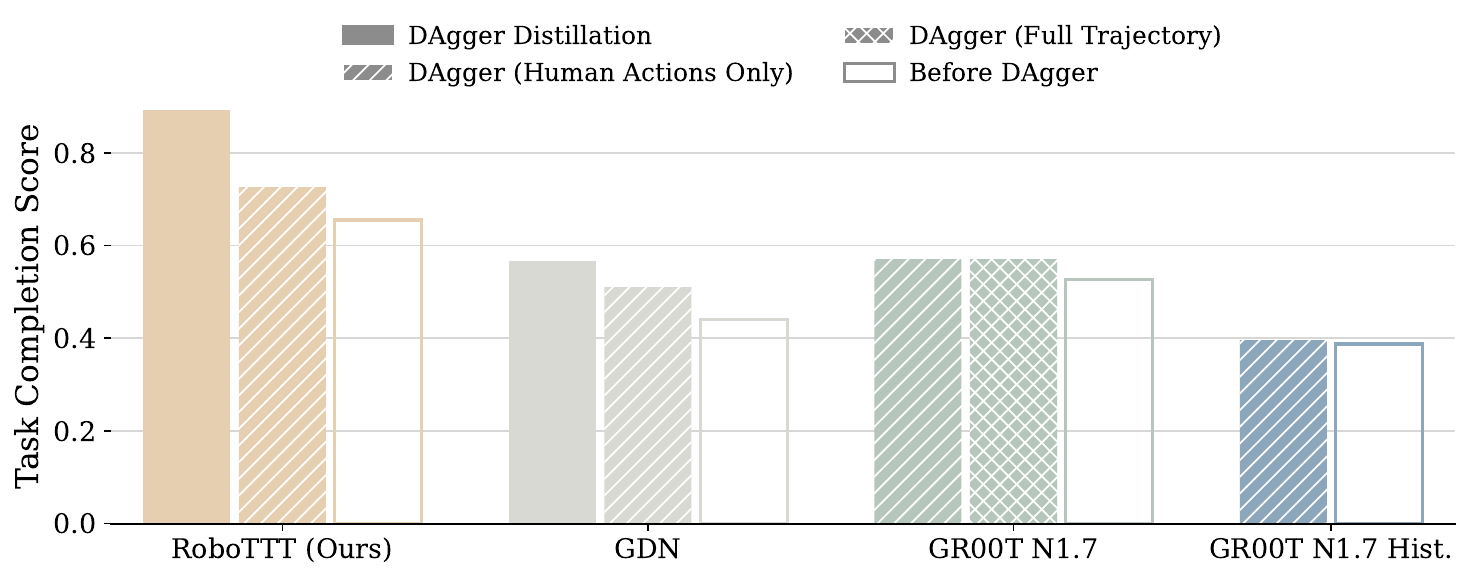}
\caption{\textbf{DAgger Distillation results on Pup Go Car.} Task completion after fine-tuning on a pool of 100 DAgger trajectories (50 collected with \acronym, 50 with \groot). \emph{DAgger Distillation} applies to the sequence models \acronym and \gdn.}
    \label{fig:dagger_distillation_results}
\end{figure}

\begin{figure}[t]
\centering
\includegraphics[width=\textwidth]{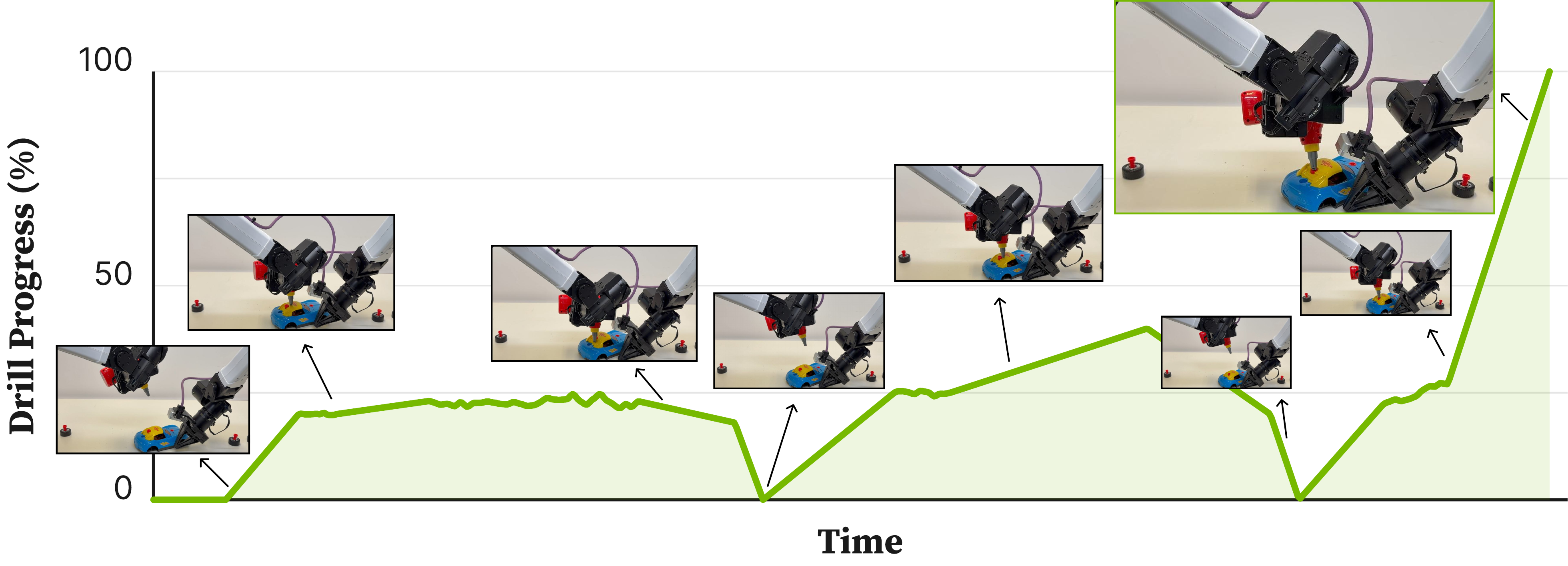}
\vspace{-0.6cm}
\caption{\textbf{On-the-fly recovery learned through DAgger Distillation.} While tightening the roof screw on Pup Go Car, \acronym misses the screw (frames 1--3), raises the arm and re-attempts, getting closer but missing again (frames 4--5), then re-adjusts once more and succeeds (frames 6--8).}
\label{fig:dagger_distillation_rollout}
\vspace{-0.2cm}
\end{figure}

\para{\acronym learns to recover on the fly and improves upon standard DAgger.}
On Pup Go Car, we study \emph{DAgger Distillation} against alternative uses of DAgger data (Fig.~\ref{fig:dagger_distillation_results}).
We collect 50 DAgger trajectories each with \acronym and \groot as the base policy, and use the pooled 100 trajectories to train all methods.
Standard DAgger, fine-tuned on human corrections alone, improves the base policies by 9\% on average across the four methods, and by 13\% on the two sequence models.
DAgger Distillation, applicable to the sequence models \acronym and \gdn, takes the full trajectory, including the suboptimal robot actions, as context while computing the imitation loss only on the human corrections (Sec.~\ref{sec:dd}).
From the same DAgger data, it yields a 33\% average improvement: 36\% for \acronym and 29\% for \gdn.
Notably, the suboptimal robot actions carry no value as imitation targets: fine-tuning \groot on the full trajectories, robot actions included, performs identically to corrections alone (57\% for both).
Their value is as context: qualitatively, most of DAgger Distillation's gain comes from recovering after wrong actions (Fig.~\ref{fig:dagger_distillation_rollout}), indicating that the failure-to-correction mappings distilled into the fast weights manifest as on-the-fly improvement during rollout.
That \gdn also improves substantially suggests DAgger Distillation applies generally to sequence-model policies, though \acronym benefits the most.

\begin{wrapfigure}{r}{0.5\textwidth}
\vspace{-10pt}
    \centering
    \includegraphics[width=0.5\textwidth]{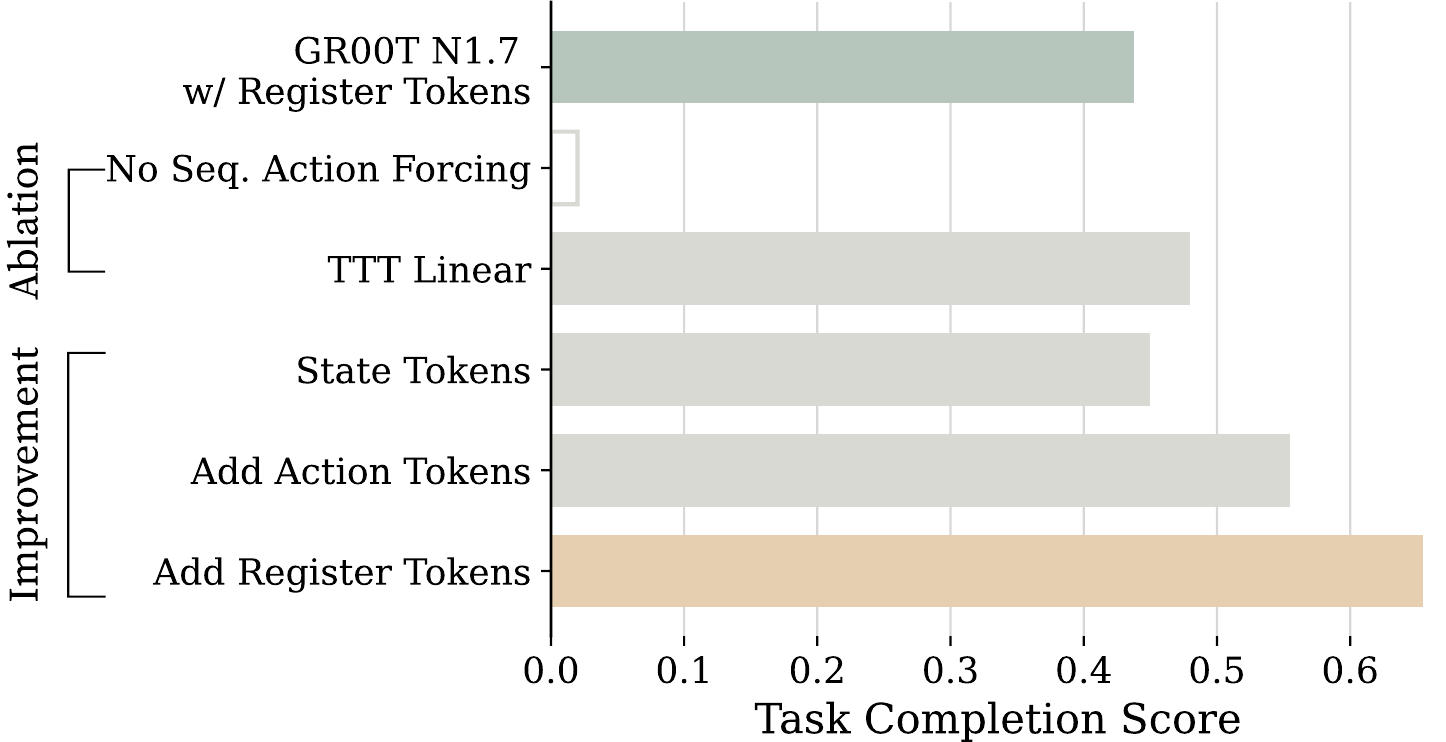}
    \caption{\textbf{Ablation results on the Pup Go Car task.} We ablate sequence action forcing and the fast-model architecture (MLP vs. linear), and trace each component added during development (state tokens, action tokens, register tokens). \groot with register tokens is included for a matched-token comparison.}
    \label{appendix:fig:ablation}
\end{wrapfigure}

\para{Design choices that matter for \acronym's performance.}
We ablate key design choices.
First, we compare against two variants: \acronym without sequence action forcing (\emph{No Seq.\ Action Forcing}) and \acronym with the fast model replaced by a linear layer (\emph{TTT Linear}).
Second, we trace the development of \acronym as a roadmap: TTT layers processing only state tokens (\emph{State Tokens}), then additionally passing action tokens through the TTT layers (\emph{+ Action Tokens}), and finally inserting learned register tokens (\emph{+ Register Tokens}), which yields the full \acronym.
Since the register tokens add capacity independent of TTT, we also augment \groot with the same number of register tokens, so the comparison isolates the TTT mechanism from the extra tokens.

As shown in Fig.~\ref{appendix:fig:ablation}, removing sequence action forcing during training significantly hurts closed-loop performance: the resulting inaccurate motions leave the robot unable to make meaningful progress. This underscores the importance of applying different noise levels across noised action chunks during sequence training. The TTT-linear variant outperforms the \groot baseline but remains suboptimal, 27\% worse than the MLP fast model, suggesting that expressive, nonlinear fast models matter most, echoing findings in vision and language modeling tasks~\citep{zhang2025test0time}.

Starting from a TTT MLP processing only state tokens, we improve performance by gradually adding tokens. Adding action tokens yields a 23\% relative improvement: aware of its past actions, the model better captures environment dynamics. Adding register tokens yields a further 18\% relative improvement. In contrast, register tokens do not help \groot. This suggests that learned register tokens are beneficial only when paired with TTT's temporal modeling, helping the model encode contextual information.

\section{Related Work}
\label{sec:related_work}

\para{Long-Context Policies}
Most state-of-the-art robot foundation models operate with single-step or short-history context.
Most Vision-Language-Action models (VLAs) and some World Action Models (WAMs) take only the current observation~\citep{nvidia2025gr00t,black2410pi0,pi05,kim2024openvla,shukor2025smolvla0,yuan2026fast0wam0,kim2026cosmos} or a few consecutive observations (typically 2 to 8)~\citep{brohan2022rt1,brohan2023rt2,team2024octo,DBLP:conf/iclr/LiuWLTCWX0025,li2025unified,ye2026world}.
Some work extends the observation window by, e.g., visually supplying movement trajectories~\citep{zheng2025tracevla}, compressing vision-language tokens~\citep{jang2025contextvla0}, predicting past actions~\citep{torne2025learning}, or caching and gating history tokens~\citep{gao2026gated}. 
These models nonetheless remain limited to a fixed context size. For longer horizons, history can be delegated to higher-level semantics such as keyframes~\citep{mark2026bpp0,sridhar2025memer0} and language~\citep{lin2025onetwovla,torne2026mem0}, but long visuomotor context remains important for capabilities such as in-context learning~\citep{duan2017one,fu2024context}, on-the-fly policy improvement~\citep{DBLP:conf/iclr/LaskinWOPSSSHFB23}, and adaptation~\citep{kumar2021rma,qi2023hand,team2023human}.

Another line of work processes the entire rollout history autoregressively~\citep{DBLP:journals/tmlr/ReedZPCNBGSKSEBREHCHVBF22,jiang2022vima,bousmalis2024robocat0,fu2024context,li2026causal}. While these models capture long-context dependencies well, they are computationally prohibitive for real-robot deployment over long horizons, since decoding latency with a KV cache grows linearly with context length. Recurrent neural network (RNN) policies~\citep{mandlekar2021matters} offer an alternative with constant inference complexity, but traditional architectures such as LSTMs~\citep{10.1162/neco.1997.9.8.1735} scale worse than their full-attention counterparts~\citep{kaplan2020scaling}. \acronym can also be viewed as an RNN policy whose recurrent states are fast weights, updated by gradient descent on the Test-Time Training (TTT) loss. Crucially, this recurrent structure lets us scale context length, and we show that \acronym consistently outperforms the best single-step and short-context baselines once context length is sufficiently scaled.

Building long-context policies requires addressing the spurious correlations introduced by history~\citep{NEURIPS2019_94701864,wen2020fighting}, where policies overfit to past actions implicitly encoded in past observations. Prior work mitigates this by summarizing context histories~\citep{sridhar2025memer0,mark2026bpp0}, introducing auxiliary objectives~\citep{torne2025learning}, or selectively bypassing context~\citep{gao2026gated}. \acronym instead addresses it through the learned TTT fast weights, which dynamically encode relevant information into the parameter space while erasing redundant features.

\textbf{\emph{Test-Time Training}} Test-Time Training (TTT)~\citep{yu2020ttt} is a paradigm in which neural networks rapidly update a small subset of their parameters, known as \emph{fast weights}, using a self-supervised objective during both training and inference, enabling continuous storage and retrieval of contextual information. Recent work has developed improved test-time optimization and online learning objectives~\citep{behrouz2024titans0,karami2025lattice0,behrouz2025it0s,zhang2025test0time}, tighter architectural integration with language and vision models~\citep{tandon2025end0to0end,feng2026inplace,han2025vit00300,zhao2026fast0weight}, and more efficient training strategies~\citep{DBLP:conf/iclr/LimZSK24,li2025tnt}. First demonstrated for language modeling~\citep{yuksekgonul2026learningdiscovertesttime}, TTT has since shown promising results in video generation~\citep{Dalal_2025_CVPR}, computer vision~\citep{han2025vit00300}, and 3D reconstruction~\citep{chen2026ttt3r3dreconstructiontesttime}. These modalities are inherently sequential, and TTT's success there suggests strong potential for robotics, where agents likewise interact with their environments in a continuous, streaming manner.

While several recent works~\citep{bai2025evolve0vla0,zhu2026ttt,liu2026fly} use the term ``test-time training'' for robotics, they do not employ fast weights, instead collecting extra data on test tasks to fine-tune the entire model. The closest setting to ours is \citet{ziakas2026vita}, which equips Vision-Language Models (VLMs) with fast weights to adapt value functions for robotic tasks. \acronym instead builds robot visuomotor policies on top of TTT layers. By scaling the training context length, we show that \acronym exhibits new capabilities such as one-shot imitation from in-context human demonstration videos and on-the-fly policy improvement, while achieving stronger closed-loop performance on long-horizon tasks.

\textbf{\emph{Robot Foundation Models}}
Recent years have seen rapid progress in robot foundation models that build on large-scale vision-language pretraining and adapt it to robotic control~\citep{brohan2022rt1,jiang2022vima,brohan2023rt2,team2024octo,kim2024openvla,black2410pi0,pi05,yang2025magma,nvidia2025gr00t,zheng2025tracevla,molmoact2}. A common paradigm initializes from pretrained VLMs and adds an action generation module mapping multimodal representations to robot actions. Existing approaches differ primarily in how actions are represented and predicted. One line formulates control as autoregressive sequence modeling, discretizing actions into tokens generated by next-token prediction~\citep{kim2024openvla,yang2025magma,pertsch2025fastefficientactiontokenization}. Another preserves continuous action spaces by coupling pretrained VLMs with diffusion or flow-matching policy heads~\citep{team2024octo,black2410pi0,nvidia2025gr00t,driess2026knowledge}, modeling multimodal action distributions more expressively. While \acronym can in principle be a plug-and-play module for any robot foundation model architecture, here we instantiate it on the flow-matching GR00T-N1.7~\citep{nvidia2025gr00t} policy, our default backbone throughout. Notably, although GR00T-N1.7 was trained with only single-step or short context, \acronym scales its context to 8K timesteps (about five minutes at 30 Hz control), and we find that capabilities such as long-context conditioning emerge only once context length is sufficiently scaled.

\section{Limitations and Conclusion}
\label{sec:conclusion}

This work has a few limitations. First, scaling training context length increases training cost; future work could adopt more recent TTT training techniques such as TNT~\citep{li2025tnt}.
Second, while we develop a principled way of integrating TTT into robot foundation models, future work might explore robotics-oriented objectives for the TTT layers (as explored for vision~\citep{han2025vit00300}).
Finally, although \acronym improves task performance substantially, it does not handle every failure mode encountered in deployment; combining it with reinforcement learning to optimize task success directly is a natural next step.

We present \acronym, a robot model and training recipe that scale the visuomotor context of robot policies to 8K timesteps.
At this context length, \acronym unlocks new robot capabilities: one-shot imitation from in-context human video demonstrations, on-the-fly policy improvement, robustness to external perturbations, and stronger closed-loop performance on multi-stage, long-horizon tasks.
At its core, \acronym integrates Test-Time Training into robot foundation models as the sequence modeling mechanism along the time dimension, and its training recipe, combining sequence action forcing with truncated backpropagation through time, makes training on long sequences tractable.
We observe for the first time that scaling pretraining context length yields steady gains in closed-loop performance, suggesting context length as a new scaling axis for robot foundation models.

\subsection*{Acknowledgments}
We thank Frederik Ebert, Letian Fu, Abhishek Gupta, Joel Jang, Hanjung Kim, Dantong Niu, Rutav Shah, You Liang Tan, Josiah Wong, Haoyu Xiong, Ruohan Zhang, Tianyuan Zhang, and Chuning Zhu for constructive discussions.
We thank Zhe Zhang and Connor Pedersen for compute cluster support.
We thank Amy Nguyen, Matin Furutan, Matin Nikoui, Mona Abbas, Lion Park, Ramanpreet Singh, Alaa Eltayeb, Dona Alhabibi, Marcelo Kulik, Nachiket Timmanagoudar, and Josiah Minor for real-robot operation, data collection, and evaluation.
We thank Tri Cao and Yuqi Xie for help with the release.
Last but not least, we thank the NVIDIA GEAR Team and the Stanford Vision and Learning Lab for their continuous support.

\bibliographystyle{plainnat}
\bibliography{references}

\clearpage
\appendix
\renewcommand{\thefigure}{A.\arabic{figure}}
\renewcommand{\theequation}{A.\arabic{equation}}
\renewcommand{\thetable}{A.\Roman{table}}

\setcounter{figure}{0}
\setcounter{equation}{0}
\setcounter{table}{0}

\section{Model Architecture, Training, and Deployment Details}
\label{appendix:sec:policy_training_details}

\subsection{Model Architecture}
We instantiate \acronym on pretrained GR00T~N1.7~\citep{nvidia2025gr00t}, which consists of the Eagle model~\citep{shi2025eagle} as the vision-language model (VLM) backbone and a Diffusion Transformer (DiT)~\citep{peebles2022scalable} as the action head. We add a TTT layer to each of its 16 DiT layers.
The original DiT has 538M parameters; each TTT layer adds roughly 10M, for 690M in total.
Each TTT layer contains a fast model, a two-layer MLP with GeLU activation~\citep{hendrycks2016gaussian}, updated by standard gradient descent at test time; more sophisticated test-time optimizers such as Muon~\citep{jordan2024muon,zhang2025test0time} are left to future work.
Following standard practice for TTT, we learn the inner test-time learning rate~\citep{yu2020ttt} on top of a constant base learning rate of 0.1. We use RoPE~\citep{su2021roformer0} for positional embeddings with $\theta_{\mathrm{rope}} = 10000$.

\subsection{Training}
\begin{figure}[h]
    \centering
    \includegraphics[width=0.5\linewidth]{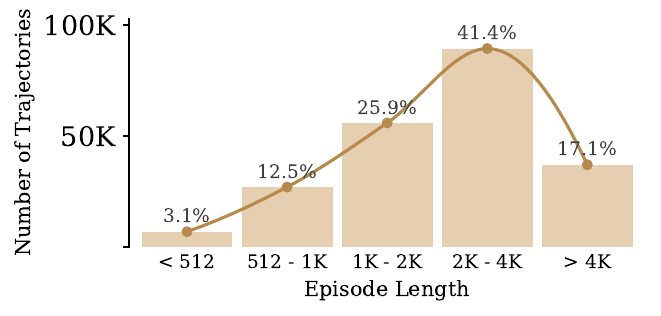}
\caption{\textbf{Trajectory length distribution of the pretraining mixture.}}
    \label{appendix:fig:data_len_dist}
\end{figure}

We pretrain on a mixture of tabletop bimanual robot data and egocentric human video data~\citep{zheng2026egoscale0}, curated to emphasize long trajectories; the trajectory length distribution is shown in Fig.~\ref{appendix:fig:data_len_dist}.
We train all models on NVIDIA GB200 GPUs, using 16 GPUs per pretraining run and 8 per task-specific post-training run.
Pretraining tunes only the newly added sequence-modeling layers (TTT or GDN) and freezes the other components of GR00T N1.7; post-training fine-tunes all parameters.
During pretraining, we use a per-device batch size of 4 (global batch size 64) for context lengths of 4K and below, and 1 (global batch size 16) above. Post-training uses a 1K context length and a per-device batch size of 1. We pretrain for 30K steps and post-train for 20K steps.
All models use the AdamW optimizer~\citep{loshchilov2017decoupled} with weight decay $1\times 10^{-5}$.
Pretraining uses the Warmup-Stable-Decay (WSD) schedule~\citep{hu2024minicpm0} with a peak learning rate of $2\times 10^{-5}$; post-training uses a cosine schedule with a peak learning rate of $5\times 10^{-5}$.

\subsection{Deployment}
We deploy trained models on the YAM bimanual tabletop robot with four RealSense D405 cameras (top, bottom, left wrist, right wrist) streaming 480p RGB observations.
Inference runs on a workstation with an NVIDIA RTX 5090 GPU, at a control frequency of 30\,Hz.

\section{Task Definition and Experiment Details}
\label{appendix:sec:experiment_settings}
This section provides task definitions and experiment details.

\subsection{Task Definition}
\label{appendix:sec:task_definition}
All tasks are scored on a [0, 1] task completion scale via task-specific rubrics, detailed below.

\begin{figure}
    \centering
    \includegraphics[width=\linewidth]{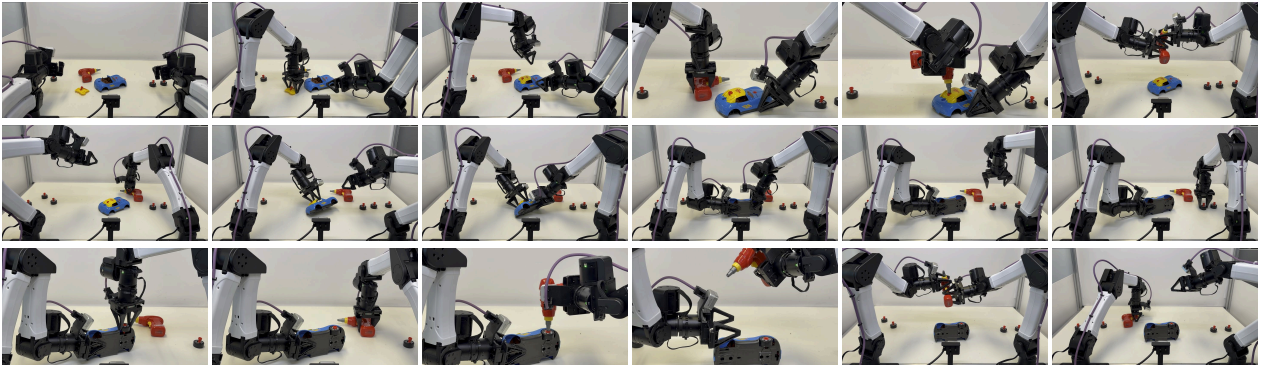}
    \caption{\textbf{The ``Pup Go Car'' task.} The robot assembles the roof and first wheel of a model car across multiple stages, including screwing, drilling, bimanual handoffs, and a car flip.}
    \label{appendix:fig:pupgo_task}
\end{figure}

\para{Pup Go Car}
As shown in Fig.~\ref{appendix:fig:pupgo_task}, the robot assembles the yellow roof and the first wheel of the model car \textit{``Pup Go Car''}. It picks up the roof, aligns the screw with the hole on the car body, and places the roof. It then grasps the drill, aligns it with the roof screw, and tightens it, before handing the drill off to the right hand. The robot next flips the car with one hand and picks up a wheel with the right hand, which inserts the wheel into the car body. It then picks up the drill again and tightens the wheel, before finally handing the drill off to the left hand.

The scoring rubric is as follows: 0.05 if the roof is picked up; 0.1 if the roof is placed on the car body; 0.25 if the roof screw is properly inserted into the car body; 0.3 if the drill is picked up; 0.35 if the drill tip contacts the screw; 0.45 if the roof screw is fully tightened; 0.5 if the drill is handed to the right hand; 0.55 if the car body is flipped and stabilized; 0.6 if the tire is picked up; 0.75 if the tire is inserted into the car body; 0.8 if the drill is picked up; 0.85 if the drill is aligned with and contacting the wheel screw; 0.9 if the wheel is fully tightened; 1.0 if the drill is handed to the left hand. We allow at most two attempts for wheel assembly.

\begin{figure}
    \centering
    \includegraphics[width=\linewidth]{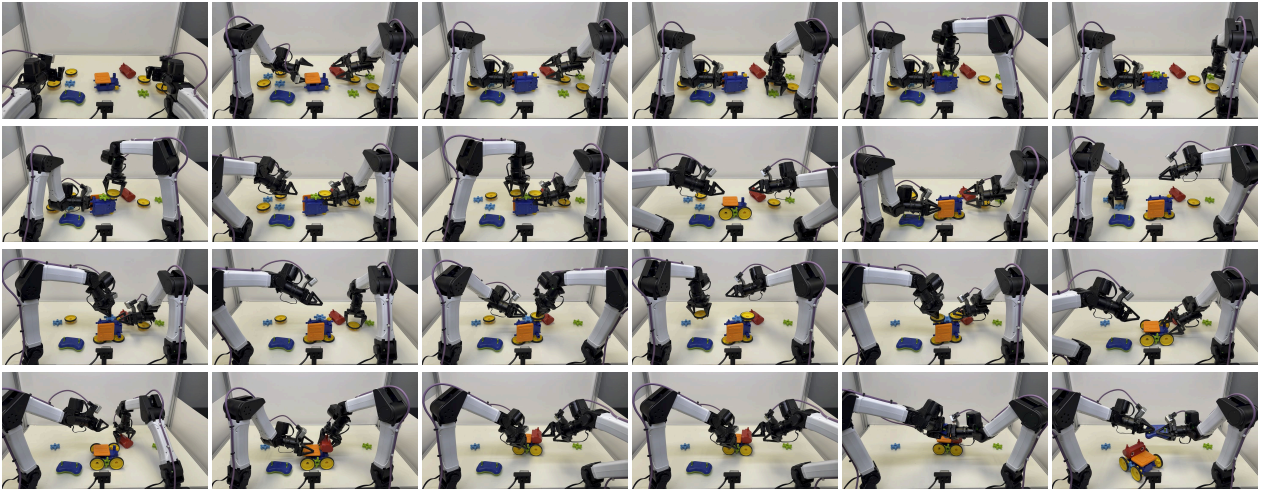}
    \caption{\textbf{The ``Gear Bot'' task.} The robot installs gears and wheels on both sides of the chassis, flips it twice, attaches the robot head, and drives it with a remote.}
    \label{appendix:fig:gear_bot_task}
\end{figure}

\para{Gear Bot}
As shown in Fig.~\ref{appendix:fig:gear_bot_task}, the robot assembles the entire toy model ``Gear Bot''. It installs one gear and two wheels on each side of the chassis, flipping the chassis twice so the shafts face upward. It then picks up the red ``robot head'' and inserts it onto the chassis. Finally, it picks up the remote control and pushes the joystick so the Gear Bot moves around. The scoring rubric is as follows: +0.1 for each chassis flip, +0.1 for each gear or wheel installation, +0.1 for ``robot head'' installation, and +0.1 if the remote control is used successfully.

\begin{figure}
    \centering
    \includegraphics[width=\linewidth]{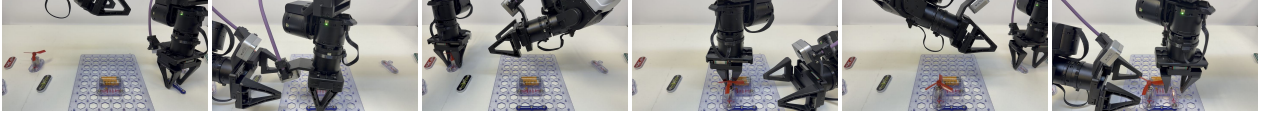}
    \caption{\textbf{The ``Circuit'' task.} The robot assembles two or three circuit components on a board and powers the circuit on when a switch or button is present.}
    \label{appendix:fig:circuit_task}
\end{figure}

\para{Circuit}
As shown in Fig.~\ref{appendix:fig:circuit_task}, the robot assembles circuit components on a board. Components include a red LED, green LED, colorful LED, lamp, motor, snap wire, press button, and switch. The robot assembles two or three pieces, following different assembly orders for the left and right components. Considering the component combinations and assembly orders, there are roughly 80 configurations. If the assembly includes a switch or press button, the robot must also turn it on after assembly. The scoring rubric is as follows. For two-piece assembly without a switch or press button, +0.5 per component installed. For two-piece assembly with a switch or press button, +0.33 per component installed and +0.33 if the circuit is turned on. For three-piece assembly without a switch or press button, +0.33 per component installed. For three-piece assembly with a switch or press button, +0.25 per component installed and +0.25 if the circuit is turned on. No partial credit is given if the assembly order is wrong.

\subsection{Experiment Details}
\para{Baselines}
For the \groot and \grootHist baselines, we use the official implementation~\citep{nvidia2025gr00t}, extended to support history-frame input for \grootHist. For the \gdn baseline, we replace each TTT layer with a Gated DeltaNet layer from the Flash Linear Attention library~\citep{yang2024fla}, keeping the layer placement, gating, and parameter count matched to \acronym.

\para{Evaluation}
We deploy trained models on YAM bimanual tabletop robots. To ensure identical initial conditions across methods, we record the initial object placements for each task and reproduce them at evaluation time. We evaluate 20 rollouts for the Pup Go Car and Circuit tasks, and 10 rollouts for the Gear Bot task and for the Circuit task under the one-shot human-video setting, owing to their substantially longer evaluation time.

\end{document}